\documentclass[journal]{IEEEtai}

\usepackage[colorlinks,urlcolor=blue,linkcolor=blue,citecolor=blue]{hyperref}

\usepackage{color,array}

\usepackage{graphicx}

\usepackage{cite}
\usepackage{amsmath,amssymb,amsfonts}
\usepackage{algorithm}
\usepackage{algpseudocode}

\usepackage{textcomp}
\usepackage{xcolor}
\usepackage{comment}
\usepackage{subcaption}  

\def\BibTeX{{\rm B\kern-.05em{\sc i\kern-.025em b}\kern-.08em
    T\kern-.1667em\lower.7ex\hbox{E}\kern-.125emX}}

\begin{document}

\title{Sequencing to Mitigate Catastrophic Forgetting in Continual Learning}

\author{Hesham G. Moussa, \IEEEmembership{Member, IEEE}, Aroosa Hameed, and Arashmid Akhavain, 
\thanks{Hesham G. Moussa (corresponding author) was with the wireless department at Huawei Technologies Canada, Kanata, Canada  (e-mail: hesham.moussa@huawei.com).}
\thanks{Aroosa Hameed was with the wireless department at Huawei Technologies Canada, Kanata, Canada  (e-mail: aroosa.hameed@h-partners.com).}
\thanks{Arashmid Akhavain was with the wireless department at Huawei Technologies Canada, Kanata, Canada  (e-mail: arashmid.akhavain@huawei.com).}
}

\makeatletter
\renewcommand{\theALG@line}{\arabic{ALG@line}}
\makeatother
\algrenewcommand\alglinenumber[1]{\scriptsize #1}

\maketitle

\maketitle

\begin{abstract}
To cope with real-world dynamics, an intelligent system
needs to incrementally acquire, update, and exploit knowledge throughout its lifetime. This ability, known as Continual learning, provides a foundation for AI systems to develop themselves adaptively. Catastrophic forgetting is a major challenge to the progress of Continual Learning approaches, where learning a new task usually results in a dramatic performance drop on previously learned ones. Many approaches have emerged to counteract the impact of CF. Most of the proposed approaches can be categorized into five classes: replay-based, regularization-based, optimization-based, representation-based, and architecture-based. In this work, we approach the problem from a different angle, specifically by considering the optimal sequencing of tasks as they are presented to the model. We investigate the role of task sequencing in mitigating CF and propose a method for determining the optimal task order. The proposed method leverages zero-shot scoring algorithms inspired by neural architecture search (NAS). Results demonstrate that intelligent task sequencing can substantially reduce CF. Moreover, when combined with traditional continual learning strategies, sequencing offers enhanced performance and robustness against forgetting. Additionally, the presented approaches can find applications in other fields, such as curriculum learning.
\end{abstract}


\section{Introduction}
\IEEEPARstart{I}{n} recent years, human intelligence has served as a benchmark for many researchers working in the field of machine learning. Today we have models that exhibit similar or even surpass human capabilities in many fields, including text (e.g., ChatGPT \cite{gpt}), games (e.g., AlphaZero \cite{alphazero}), and object recognition (e.g., YOLOv11 \cite{yolo}). While these results are impressive, the underlying models in almost all these fields are typically trained to be static and incapable of adapting their performance and functionality over time. This is attributed to the fact that the training process is usually done over a fixed set of data samples that is often designed to train a model to achieve a certain task. If the dataset changes or evolves, or the task to be learned changes, the training process needs to be restarted (often from scratch). However, as we live in such a dynamic world, data is continuously being generated, and knowledge is ever-evolving; thus, retraining a model every time new data becomes available or every time a model needs to learn a new task might not be feasible, as it might require prohibitive storage, computing, and power resources. 

Such static intelligence is different than human intelligence. Humans are capable of sequentially and continually acquiring knowledge and learning new skills, allowing them to perform multiple tasks simultaneously with ease. A unique aspect to highlight here is that humans can easily integrate new learned knowledge and skills without erasing or interfering with previously learned capabilities. For instance, a child who masters walking can then acquire the ability to ride a bicycle without compromising their ability to walk. Also, students can concurrently learn multiple subjects at school without mixing distinct conceptual frameworks, and do well on all tests.

Replicating the dynamic learning abilities of humans in the realm of machines and artificial intelligence (AI) is a focus of a relatively new field called continual learning (CL) \cite{surveycl}. CL explores the problem of learning tasks from a stream of data sequentially and continuously, leading to a gradual expansion of the model's acquired knowledge and enabling multi-task handling without fully re-training the model. In the literature, CL is also known as sequential learning, incremental learning, and life-long learning. 

A major challenge in CL is catastrophic forgetting (CF) \cite{49-2}, also called catastrophic interference (CI). CF occurs when a model abruptly loses previously acquired knowledge while being trained on new tasks. This happens because model training updates the network’s weights, typically via optimization methods like stochastic gradient descent (SGD), to perform well on the new task. If the optimal solutions for the new and old tasks differ, the model may overwrite previous knowledge, resulting in a sharp drop in performance on earlier tasks. It should be noted that recent studies have shown that even large language models (LLMs) seem to suffer from CF, especially during the phase of fine-tuning \cite{LLmCF2, llmCF1}.

When training a model using CL, different tasks are presented sequentially, and the objective is to learn the stream of tasks without catastrophically forgetting. Several CL methods have been proposed in recent years to address different aspects that might lead to varying degrees CF. Conceptually, these methods can be categorized into five groups based on the aspect of CF they target. These categories include: Replay-based methods, Regularization-based methods, Optimization-based methods, Representation-based methods, and Architecture-based methods. While each category has advantages and disadvantages, it has been shown that a mixture of them often results in the best mitigation results.

In this work, we explore a different dimension of the CL problem compared to the aforementioned five categories. To be exact, we investigate the impact of the sequence of tasks on the CF performance. We attempt to answer four fundamental questions: i) Does the sequence by which the tasks are fed to the model matter?; ii) is there a way to choose an optimal sequence of tasks to learn while avoiding CF?; iii) does the distribution of the data in the different tasks impact the overall performance?; and iv) can mixing optimal sequencing and CF approaches improve the performance of CL?. As such, the contributions of this work are fivefold:
\begin{itemize}
    \item We investigate the impact of task sequencing on the performance of CL and draw valuable conclusions;
    \item We formulate the CF problem as a stochastic optimization problem and determine the challenges associated with finding a realistic and efficient solution;
    \item We propose a method that uses architectural evaluation to determine the best next node in the optimal sequence as a potential first step towards solving the CF optimization problem. The proposed method is tested under semi-i.i.d. and non-i.i.d. scenarios. The setup assumes a centralized controller node that oversees the CL process;
    \item We propose a modification to the architectural evaluation method to make it more suitable for the case of non-i.i.d sequence of tasks;
    \item We study the potential advantages that might arise when the proposed optimal sequencing method is mixed with CL methods. We specifically combine our method with elastic weight consolidation (EWC) \cite{49-2}, a weight regularization-based approach.
\end{itemize}

The remainder of this paper is organized as follows. Section 2 provides background on CL and a high-level review of approaches in the literature that aim to mitigate CF. Section 3 describes the system model considered in this work. In Section 4, the CF problem in CL is formulated as a stochastic optimization problem. Section 5 reports establishes a baseline relationship between sequencing and CF. Section 6 introduces the proposed method for optimal next-node decision making, along with the necessary modifications for handling different data distributions. Section 7 outlines the experimental setup and presents the optimal sequencing results across various scenarios. Finally, Section 8 concludes the paper with discussions and directions for future work

 Note. To limit the scope, our reported results are based on class-based CL scenarios. However, the results could be extended to other types of CL scenarios.

\section{Related Work}
\subsection{Background on Continual Learning}
\begin{figure}[]
\centering
 {\includegraphics[width=0.40\textwidth, height=8cm]{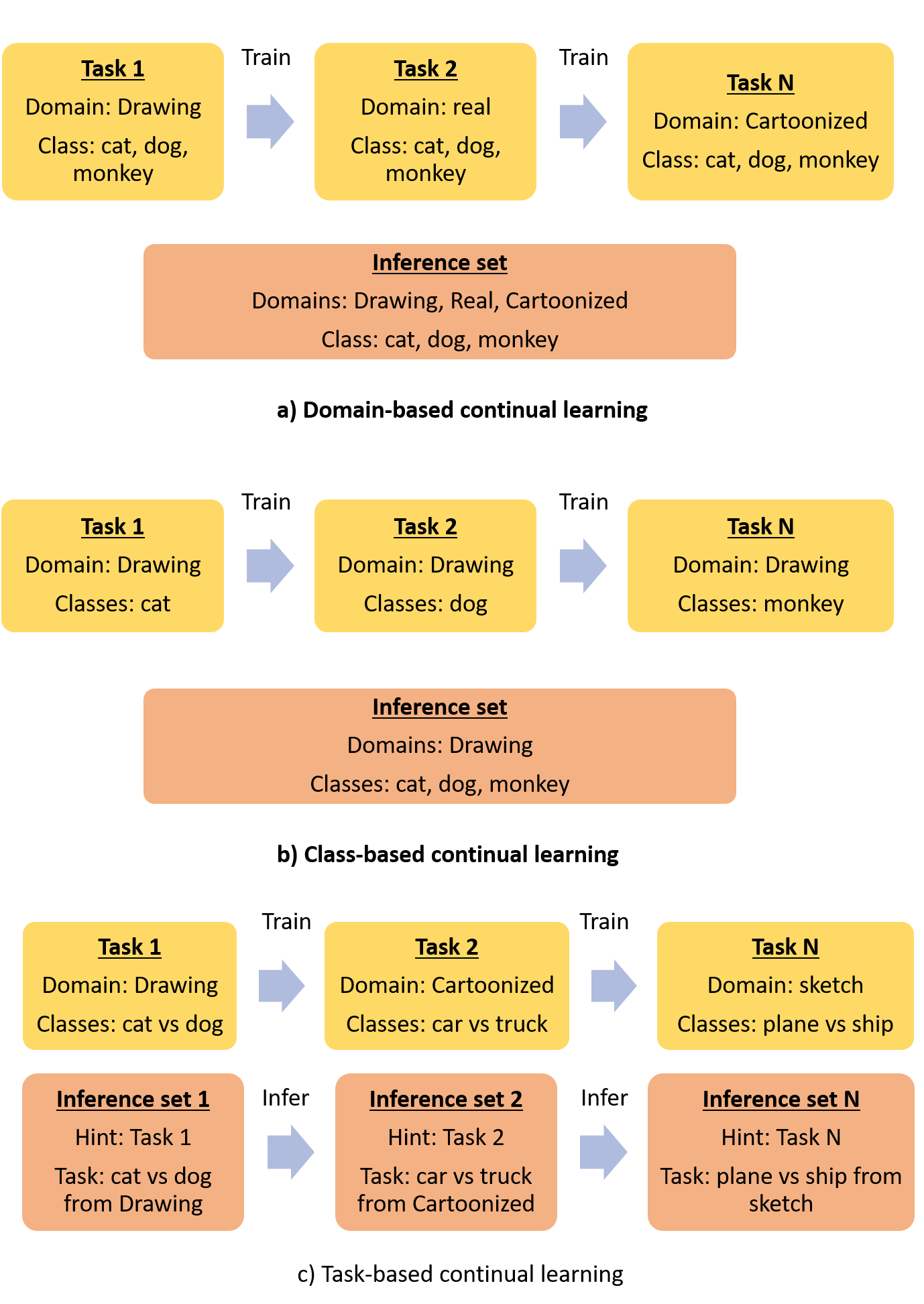}}
 \caption{Different types of Continual Learning Scenarios}
 \label{fig:CL_scenarios}
\end{figure}

CL deals with a range of problems depending on the differences between the tasks being presented to the model. At its core, and as shown in Figure \ref{fig:CL_scenarios}, differences between tasks can be categorized into three types, domain-based, class-based, and task-based \cite{surveycl}.

Domain-based CL focuses on scenarios where the change is in the domain, conceptual drift, or distribution shift (e.g., a change in the environment of an image, an image taken during the day versus one taken at night), but the model sequentially encounters the same set of classes. For example, as in Figure \ref{fig:CL_scenarios}(a), the model first learns to classify animals (cats, dogs, monkeys) in drawings, then learns to recognize these same animals in real images (e.g., photographs), then in cartoonized images, and so on. Essentially, the classes remain constant, but the visual style, context, or other distributional properties change. The focus here is on adapting to distribution shifts while maintaining performance across all previously encountered domains.

In class-based CL, in each task of training, a new subset of classes, not available in other tasks, is introduced. For example, the model might first learn to classify cats and dogs, then learns birds and fish, then cars and trucks, and so on. The objective is to continuously learn the new classes introduced in each task while preserving the model's ability to recognize all previously learned ones. This has to be done without having access to training data from previous tasks (Figure \ref{fig:CL_scenarios}(b)).

In task-based CL, a model sequentially learns multiple tasks, and at test time, the task identity is provided to the model in a hint-like fashion. For example, as in Figure \ref{fig:CL_scenarios}(c), the model learns Task 1 (classifying cats vs dogs from the domain of Drawings), then Task 2 (classifying cars vs trucks from the domain of Cartoonized), and then Task 3 (classifying planes vs ships from the domain of sketch), with each task potentially having its own set of classes, domains, and/or objectives. The key distinguishing feature is that during inference, the model knows which task it is being asked to perform. For example, if shown an image, it is told, "this is from Task 2," so it only needs to distinguish between cars and trucks rather than among all six classes across all tasks and domains. This task-identity information significantly simplifies the problem compared to class-based CL, as the model can use separate output heads or decision boundaries for each task. 

\subsection{CL methods to mitigate Catastrophic forgetting}
Several CL methods have been proposed in recent years to address different aspects of machine learning, with varying degrees of success in mitigating CF. Conceptually, these methods can be categorized into five groups based on the aspect they target, namely, Replay-based methods, Regularization-based methods, Optimization-based methods, Representation-based methods, and Architecture-based methods.

In replay-based methods, samples from learned tasks are stored either in raw form or as representative generated pseudo-samples, and are replayed during the learning of a new task. Replay-based methods include rehearsal and constraint optimization approaches. In rehearsal, the replay samples are combined with the new task’s training data to train the model \cite{16, 49, 50, 51}. Constraint optimization methods use replay samples to limit changes in model weights during new task learning by adding a constraint to the optimization problem, which is a function of the replay samples \cite{6, 55}.

In regularization-based methods, the loss function is modified by adding a regularization term to preserve previously learned knowledge as new tasks are learned. They are divided into data-based approaches, such as methods that employ techniques from the field of knowledge distillation \cite{118,16,119}, and weight-based approaches, where model weights are constrained during new task learning \cite{49-2,108,109}.

Optimization-based methods are explicitly designed for manipulating the optimization process during training. Examples include gradient projection based on old tasks \cite{67}, meta-learning within the inner training loop, and exploiting mode connectivity in the loss landscape \cite{86,113}.

Representation-based methods mitigate forgetting by preserving the learned feature space so that representations of old tasks remain stable while new tasks are acquired. Unlike methods that only constrain weights or store samples, they focus on learning task-invariant and transferable representations. For instance, contrastive and self-supervised learning modules encourage features that are resistant to distribution shifts across augmentations or modalities \cite{coral}. Other approaches maintain the geometry of the feature space by reconstructing past representations or replaying pseudo-features using generative models \cite{niu2024}. Recent work also shows that continual training of multi-modal models (e.g., CLIP) can cause spatial disorder in the representation space and proposes strategies such as maintaining off-diagonal information in contrastive similarity matrices to preserve inter- and intra-modal consistency \cite{clip}.

Last, in architecture-based methods, task-specific parameters are introduced through architectural modifications. Strategies include assigning dedicated parameters to each task (parameter allocation) \cite{221}, constructing task-adaptive sub-modules or sub-networks (modular networks) \cite{244}, adding new layers or nodes (model expansion) \cite{227}, and decomposing the model into task-shared and task-specific components (model decomposition) \cite{238}. The literature also includes hybrid approaches that combine multiple strategies \cite{228}.

\subsection{Sequencing impact on catastrophic forgetting}

As has been shown in the aforementioned section, the dominant approaches to mitigating CF under CL training paradigm fall into five principal families, each addressing training dynamics or architectural design rather than the order in which tasks are presented. Consequently, sequencing has historically been absent from standard CL taxonomies when discussing CF mitigation methods.

Although sequencing is not well studied under CL, it is, in fact, the central objective of an alternative yet related field of research known as curriculum learning (CuL) \cite{CuL1}. CuL refers to the strategy of training models by presenting data in a meaningful sequence, beginning with simpler examples and gradually progressing to more complex ones. This sequencing has been shown to have a considerable impact on model performance. Unlike random data shuffling, carefully designed curricula leverage the ordering of samples to stabilize optimization, accelerate convergence, and improve generalization across tasks. It is important to mention that some studies suggest that the right curricula and ordering of the samples can indirectly mitigate CF, as it reinforces previously acquired knowledge before introducing higher-complexity tasks \cite{CuL2}.

However, the challenge in CuL lies in defining what constitutes “easy” versus “hard” samples and determining the pacing function for introducing complexity. Thus, while a particularly important benefit of sequencing in CuL is its potential positive impact on CF mitigation, this effect is essentially a byproduct rather than the primary aim of the CuL field. This distinction constitutes the main difference between CuL sequencing solutions and CL solutions: sequencing solutions in CuL are designed to improve convergence speed, stability, and generalization within a single task or dataset, whereas sequencing methods in the realm of CL should be explicitly designed with the intention of enabling models to learn multiple tasks sequentially over time while retaining previously acquired knowledge and avoiding CF.

Nonetheless, evidence from CuL, that sequencing can help mitigate CF, provides strong motivation for current efforts to explore task sequencing methods under CL paradigm. For instance, in a recent study that considered sequence transferability and task order selection in CL, it was demonstrated that task order is not a variable that can be neglected, but is rather a critical determinant of performance \cite{CuL3}. The study shows that certain task sequences facilitate positive transfer, improve retention, and reduce CF, while poorly chosen sequences aggravate forgetting. In another work, the impact of sequencing on CF was also confirmed by Bell et. el \cite{bell}, where they demonstrated that order does affect CL performance and they have presented a simple approach for choosing the right order.

Such findings highlight sequencing as a central design variable in CL systems, rather than a peripheral consideration. However, choosing the right sequence that would mitigate CF in CL remains understudied. Motivated by these insights, the present work advances sequencing as an explicit CL methodology, where we attempt to propose sequencing methods and task‑ordering strategies tailored to CF mitigation. The developed methods focus on being lightweight and assume the data is distributed across multiple nodes rather than in a central location, as will be discussed in the next few sections.

\section{Sequenced CL system model}

\begin{figure}[]
\centering
 {\includegraphics[width=0.45\textwidth]{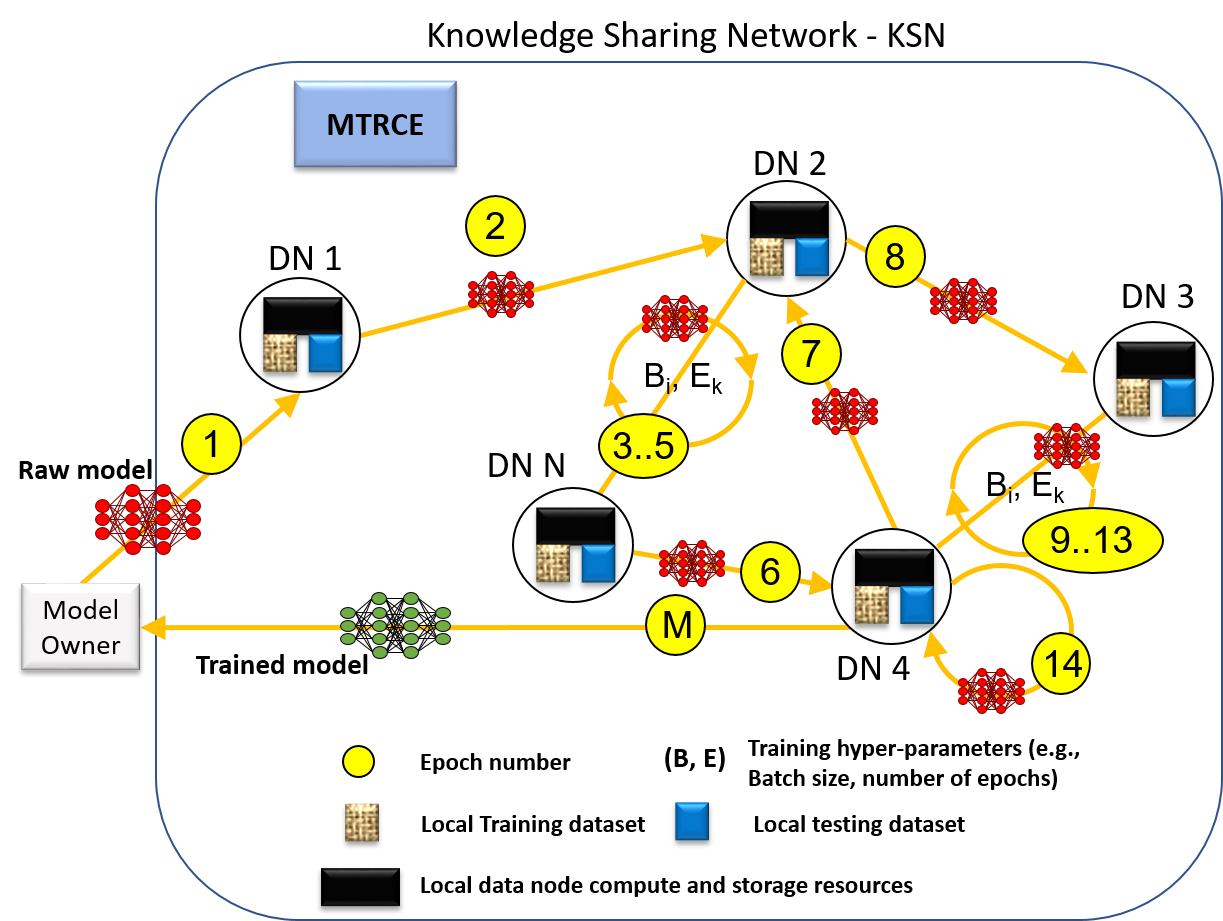}}
 \caption{System model considered consisting of a knowledge sharing network (KSN), which is composed of a number of networked data nodes (DNs) and a model training compute engine unit (MTRCE) hosted at a centralized controller}
 \label{fig:system_model}
\end{figure}

Consider a knowledge-sharing network (KSN), shown in Figure \ref{fig:system_model}, consisting of N data nodes (DNs). The DNs are assumed to have communication capability, allowing them to connect in an ad hoc manner. The DNs may locally host computing and storage resources or have access to the resources remotely, enabling them to conduct model training.

A global training dataset ($D_{train} = \{(x_i, y_i)\}_{i=1}^{I}$) is distributed across the N DNs such that each holds a portion denoted by $D_{train}^n = \{(x_k, y_k)\}_{k=1}^{K}$, where $\sum_n^N D_{train}^n = D_{train}$ (overlapping of samples is permissible), and $K \leq I$. Each portion of training data is equivalent to a training task in the realm of CL. Additionally, a global testing dataset, denoted by $D_{test}$, is present in the network. This testing dataset manifests itself in either a centralized or decentralized manner. In the decentralized embodiment, $D_{test} = \{(x_t, y_t)\}_{t=0}^T$ is created as an aggregation of sub-test-datasets, where each sub-test-dataset is held by a DN in the network. The $n^{th}$ sub-test-dataset is created by the respective $n^{th}$ DN by spliting its local portion of training dataset, $D_{train}^n$, into a sub-training ($D_{train}^{n,sub}$) and a sub-testing ($D_{test}^{n,sub}$) datasets such that $D_{train}^{n,sub} \cup D_{test}^{n,sub} = D_{train}^{n}$ and $D_{train}^{n,sub} \cap D_{test}^{n,sub} = \emptyset$. The sub-test-datasets are assumed to stay local at the DN, and to conduct a testing epoch, the model will need to conduct some form of continual testing (out of scope of this work).  On the other hand, in the case of the testing dataset manifesting itself in a centralized manner, which is the scenario that is considered in this work, a copy of each sub-test-dataset is collected from each DN and placed in a centralized location, accessible to all DNs. A testing epoch in this case would involve a single visit to this central location and running a model testing and evaluation logic.

Machine learning models requiring training are submitted to the KSN, which executes a series of steps to complete the training process using CL by moving the model between the DNs. To do so, the model visits one DN at a time to train on its local data. Once the model completes training on a DN, the next training DN (training hop) is chosen, and the model is moved towards it. Multiple visits to the same DN are allowed either sequentially or sporadically throughout training. Also, training hyperparameters may change from one DN to the other. Training runs in this fashion until one or more predefined stoppage criteria, such as a certain number of epochs, a target accuracy performance, a training time...etc, are met. It should be noted that this setup is inspired by the work presented in \cite{hesham}.

Furthermore, it is assumed that a central entity, referred to herein as the model training route compute engine (MTRCE), is responsible for making decisions that guide the movement of the model between DNs. The MTRCE houses algorithms used to compute the next DN in the sequence. It is also responsible for optimizing the training hyperparameters per DN while considering the model's evolution. The MTRCE is also responsible for housing the global $D_{test}$ dataset to evaluate the model performance after each DN visit; the objective is to regularly assess the final accuracy of the model on the overall distributed data. Once the model is trained (i.e., the stoppage criteria are satisfied), the trained model is returned to the model owner and exits the KSN.

\section{problem Formulation}

In its simplest form, the problem can be envisioned as finding the best next hop given the current state. The current state depends on the partial sequence that the model has visited so far. As such, assume that as the model moves from one DN to the next, this constitutes a step (epoch of training). The objective is to predict the best next DN the model should visit next that: i) maximizes the overall positive performance on the global datasets; and ii) reduces catastrophic forgetting performance on the individual datasets. Thus, the problem can be formulated as follows.

Let $v_h$ denote the predicted DN for step number $h$ in the CL training sequence, where $v_h \in\{1,\dots,N\}$. Also, let $S^{h-1} = \{v_1, v_2, v_3,...,v_{h-1}\}$ denote the stochastic partial sequence of visited DNs/training datasets until step $h-1$, where $v_i\in\{1,\dots,N\}$ for $i \in\{1,\dots,h-1\}$, and the corresponding probability of the sequence is given by $P^{h-1}$. It is important to notice that there are many possible partial sequences up until step $h-1$ that the model could have followed. As such, we are dealing with a conditional optimization problem where the predicted next DN depends on the preceding sequence of DNs visited, as well as the performance of the model after training on that partial preceding sequence. 

Denote the model weights after training on a certain sequence of $t$ DNs by $\Theta_{t}(S^t)$ and the corresponding performance on a test dataset $D$ by $f_{\Theta_{t}(S^t)}(D)$. Now, for compactness, we define $M(\Theta_t(S^t)) = f_{\Theta_t(S^t)}(D_{test}) $ as the model performance on the global dataset after training it on the sequence $S^t$. Thus, the one-step global performance change as the model moves from the current DN to the next DN in step $h$ can be defined as:
\begin{equation}
\Delta M(v_h|S^{h-1}) = M(\Theta_h(S^{h-1} \oplus v_h)) - M(\Theta_{h-1}(S^{h-1})),
\end{equation}
where $S^{h-1} \oplus v$ appends $v$ to the sequence $S^{h-1}$ to create the sequence $S^h$.

Moreover, a condition on the set of possible DNs from which the next DN can be chosen can be considered. To do so, let the set of DNs from which the next DN at step $h$ can be selected be denoted by $C_h \subset \{1,\dots,N\}$. $C_h$ can be designed such that different groups of DNs be considered. For instance, $C_h$ could denote all N DNs in the network ($C_h = \{1,\dots,N\}$), all the N DNs in the network except the current DN that the model has trained on in step $h-1$ ($C_h = \{\{1,\dots,N\} - \{v_{h-1}\}\}$), all the DNs that have not been chosen so far ($C_h = \{\{1,\dots,N\} - \{S^{h-1}\}\}$) , or alternate between the above options depending on the current state of the model performance and/or intended objective.

Furthermore, one way to consider the factor of catastrophic forgetting is by placing a threshold on the maximum loss of performance allowed to occur on previously learned tasks. Accordingly, let $\epsilon<0$ denote this maximum allowable catastrophic forgetting factor. Also, let $\Delta W_h^j$ denote the one-step loss in model performance on the $j^{th}$ previously learned task in the sequence $S^{h-1}$, which can be defined as 
\begin{equation}
    \Delta W_j = f_{\Theta_{h}(S^h)}(D_{test}^{v_j,sub}) - f_{\Theta_{h-1}(S^{h-1})}(D_{test}^{v_j,sub})  
\end{equation}
where $f_{\Theta_{h}(S^h)}(D_{test}^{v_j,sub})$ is the performance of the model that was trained on the datasets determined in the sequence $S^h = S^{h-1} \oplus v_h$ on the dataset associated with the $j^{th}$ DN in the sequence $S^{h-1}$.

Therefore, the condition on CF that needs to be maintained as the model moves from the current DN to the next in step $h$ and trains on the associated data can be defined as
\begin{equation}
    \min{ \{\Delta W_h^j\}_{v_j \in S^{h-1}}} \geq \epsilon, \quad for \quad j = {1,\dots, h-1}
\end{equation}
 
Accordingly, the optimization problem of selecting the next DN at step $h$ that shall maximize the expected post-step global performance of the model conditioned on the event that, up until step $h-1$, the model has followed the sequence defined by $S^{h-1}$, which has the probability $P^{h-1}$, and constrained by a maximum allowable CF for tasks associated with previously visited DNs in the sequence, can be defined as
\begin{equation}
\begin{aligned}[b]
    v_h^*  &= \underset{v \in C_h}{\arg\max} \quad   \mathbb{E}_{S^{h-1} \sim P^{h-1} } \bigg[ \mathbb{E} \bigg[M(\Theta_h(S^{h-1} \oplus v))\big | S^{h-1}\bigg]\bigg] \\
    \textrm{s.t.}\\
    &\min{ \{\Delta W_h^j\}_{v_j \in S^{h-1}}} \geq \epsilon, \quad for \quad j = {1,\dots, h-1}
    \end{aligned}
\end{equation}

An equivalent formulation is to maximize the rate of change from one step to the next, which can be defined as

\begin{equation}
\begin{aligned}[b]
    v_h^*  & = \underset{v \in C_h}{\arg\max} \quad   \mathbb{E}_{S^{h-1} \sim P^{h-1} } \bigg[ \mathbb{E} \bigg[\Delta M(v|S^{h-1})\big | S^{h-1}\bigg]\bigg]\\
    \textrm{s.t.}\\
    &\min{ \{\Delta W_h^j\}_{v_j \in S^{h-1}}} \geq \epsilon, \quad for \quad j = {1,\dots, h-1}
    \end{aligned}
\end{equation}

It follows from the formulation above that a naïve, brute‑force solution requires computing $\Theta_h$ and $M(\Theta_h)$ for every candidate $v_h \in C_h$. Concretely, this entails training the model on the local data associated with each candidate DN to produce $ L = |C_h|$ distinct model instances, evaluating the expected performance of each instance on the global dataset and on each testing dataset associated with every DN visited across the sequence $S^{h-1}$, and selecting the candidate that maximizes the expected metric while minimizing CF. Such an approach is computationally prohibitive: it multiplies training and evaluation cost by $L$, incurs substantial time-to-conclusion periods, and wastes resources on candidates that ultimately will not be selected. It is therefore desirable to avoid unnecessary training runs by estimating or otherwise screening candidates before committing to full model updates. The method proposed in this paper is designed in an attempt to provide precisely that capability.

It is worth noting that in an alternative formulation, the prediction could be extended to decide on a horizon of length $J$, i.e., predicting the best sequence of the next $J$ DNs. However, because the model changes after each visited DN, a sequence chosen now can become suboptimal as soon as the first DN is visited, forcing recomputation at every step and eroding any computational savings. More fundamentally, it is hard to predict how training on a DN’s local data based on the computed sequence will affect global performance without actually performing the full training, since the outcome depends strongly on the model state after each training step in the sequence. A similar conclusion also applies when estimating the maximum expected CF resulting from the sequencing decision. These considerations motivate a simpler step‑wise selection strategy that is efficient, which leverages inexpensive proxies to approximate $\Delta M(v|S^{h-1})$ and the expected CF thereby avoiding repeated full retraining episodes.

\section{Baseline: random sequencing}
In this section, we investigate the impact of sequencing on CL, and we attempt to establish a baseline. To do so, the KSN is initialized with six different DNs, each carrying a portion of a global dataset. We chose the clipart domain dataset from the DomainNet dataset for this section. DomainNet dataset is a large multi-domain dataset consisting of six domains, each having at least 45k data samples of 345 classes \cite{domainnet}. This dataset is well-suited for testing different scenarios of CL and hence we will be using it to generate all the results presented in this paper. 

To establish a baseline and draw some meaningful conclusions on the impact of sequencing on CF, data from the clipart domain is first divided into six portions. The data is divided in a manner that achieves a non-i.i.d. structure. To do so, the data is divided into disjoint class-wise sub-datasets $D_i$, each having a $k_i$ unique set of classes (i.e., $k_i \cap k_j = \emptyset \quad \forall \quad i \neq j$ ), where each $D_i$ contains $30\leq |k_i| \leq 60 $ classes, and each sub-dataset has a total number of samples of $6k\leq |D_i| \leq 10k$ samples for $i = \{1,2,..,6\}$. By doing so, we would be creating a class-based CL scenario.

\begin{figure}[]
\centering
 {\includegraphics[width=0.45\textwidth, height=8cm]{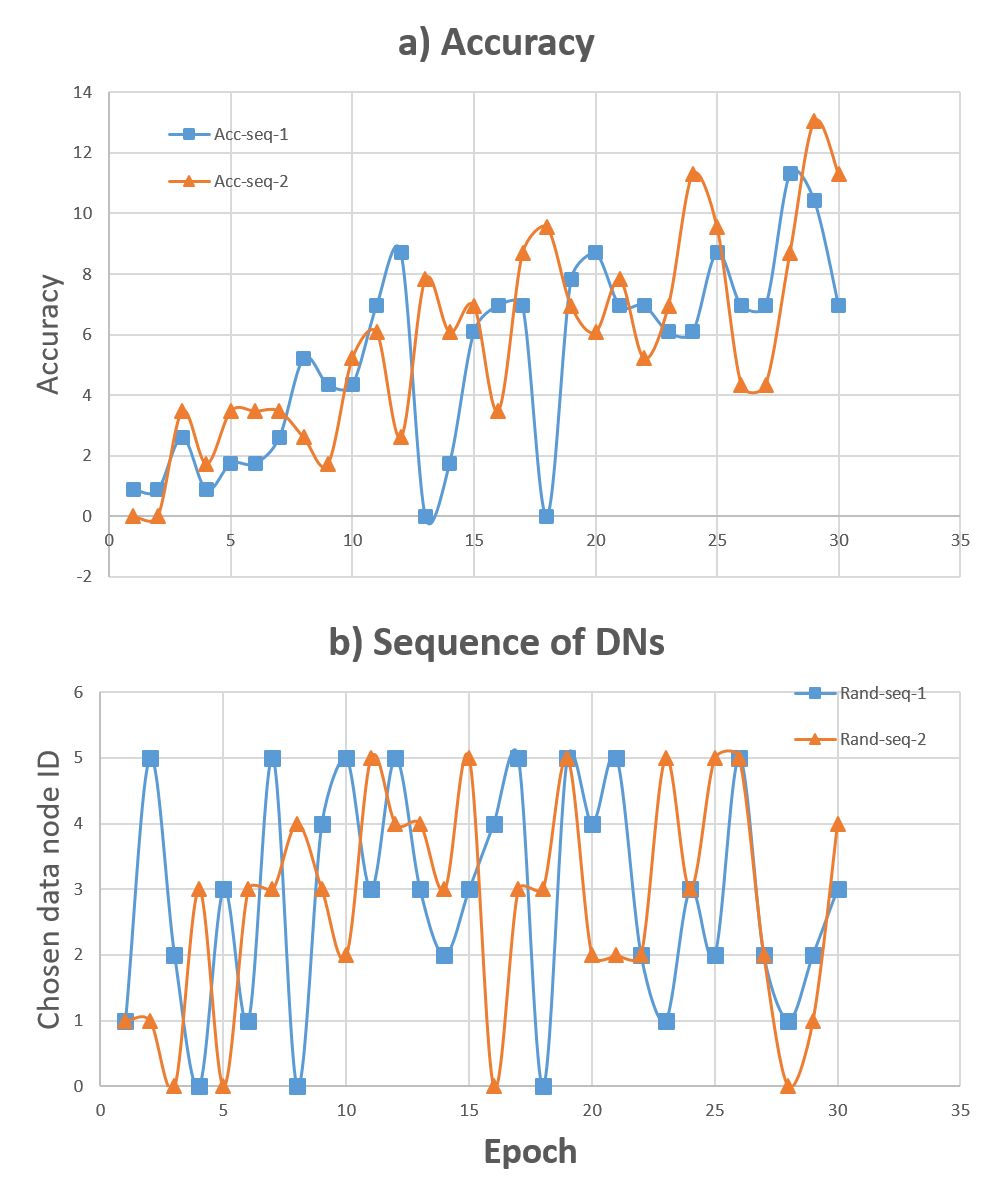}}
 \caption{Baseline results: a) step-by-step accuracy of the model as it moves from one DN to the next as per a predefined random sequence, b) Sample of the two random sequences of training followed by the model}
 \label{fig:baseline}
\end{figure}

As for the model, we used a raw non-pretrained version of ResNet18 ML model \cite{resnet} to learn the clipart dataset sequentially. The model is moved between DNs according to a randomly generated sequence. For this experiment, we generate two random sequences of 30 epochs (hops) each for the model to follow and compare the accuracy results as shown in Figure \ref{fig:baseline}. The results show an average of four runs per sequence. Once the model lands on one of the DNs, it trains for a full epoch (i.e., it consumes the full samples of the dataset hosted at the respective DN). Training hyperparameters are fixed for the 30 epochs and are kept the same for the two sequences. The reported accuracy in Figure \ref{fig:baseline}(a) is computed by testing the model on the full clipart test dataset, which contains samples from all 345 classes \cite{domainnet}.  

As can be seen, changing the sequence of DNs that the model visits impacts the performance. One sequence resulted in a final average accuracy of 11\% while the other ended up with 7\% accuracy. However, it is clear from the irregular graphs of accuracy and low performance that CF took place as the model moved from one DN to the next. However, the impact of CF was not consistent. For instance, CF was more severe when the model moved from DN 3 to DN 5 and then back to DN 3 (epochs 11-13, random sequence 1), compared to moving from DN 3 to DN 4 and then back to DN 3 (epochs 7-9, random sequence 2). As such, we conclude that choosing the right sequence of tasks may impact CF, which confirms findings from the literature as well as motivates us to explore possible ways of choosing the optimal next DN. It should be noted that, for this experiment, no CL method to mitigate CF has been applied; thus, any apparent improvement in CL performance may be attributed directly to sequencing. 

\section{Proposed Sequencing Methods} \label{sequencing_framework}
Motivated by the above discussion, it becomes evident that identifying the optimal next DN in a sequence (one that maximizes model performance while simultaneously minimizing CF) has an impact on the performance of CL; yet it is clearly a computationally challenging problem. Traditional approaches often require extensive training and evaluation, which can be prohibitively expensive in terms of both time and resources. Consequently, it is desirable to explore methods that can predict the expected performance of a model on a given dataset without the need for full training. 

One notable work in this field was presented by Bell et al. in \cite{bell}. In this study, the authors investigated how task ordering impacts continual learning performance, demonstrating that reordering tasks can either exacerbate or mitigate CF. Drawing inspiration from curriculum learning, they linked task sequencing to performance outcomes and introduced a simple yet effective method that computes the distance between tasks, defined as the curvature of a gradient step from one task toward another. This metric enables the design of task orders that reduce forgetting. The proposed method requires training multiple versions of the model, with each version trained on one of the available tasks. Each model is then tested on a sample dataset from the other nodes. Subsequently, a matrix of measured curvatures is constructed and used to select the next best node. However, this process must be repeated after each step, which demands considerable computational resources and takes time.

To this end, and in an attempt to find alternative lightweight methods to speed up the selection of the next DN, we draw inspiration from the field of Neural Architecture Search (NAS). In particular, recent advances in zero-shot NAS have attracted significant attention due to their ability to estimate model performance without training, thereby accelerating the search for architectures that best fit a given dataset \cite{NAS1, NAS2}. These methods leverage proxy signals, such as gradient norms, weight distributions, or information-theoretic measures, to approximate how well a candidate architecture is expected to perform; thus, reducing the computational burden of exhaustive training and evaluation \cite{NAS-survery}.

It is important to note, however, that the objectives of zero-shot NAS differ fundamentally from those of sequencing in CL. In NAS, the problem is framed as selecting the best architecture from a large search space for a single dataset. In contrast, CL sequencing involves a fixed model that must be applied across a sequence of datasets, where the challenge lies in determining the optimal ordering of datasets to balance knowledge acquisition and retention. In other words, NAS explores a multitude of models for one dataset, whereas CL sequencing explores a number of datasets for one model, effectively inverting the problem structure.

Despite this difference, the methodological insights from zero-shot NAS are highly relevant. Inspired by this line of work, we propose incorporating zero-shot NAS algorithms into a framework designed to approximate $\Delta M(v|S^{h-1})$. By leveraging zero-shot predictors, we can efficiently estimate the relative utility of each candidate dataset among the $N$ possible options, and subsequently select the dataset that maximizes the expected performance gain while mitigating CF.

This approach establishes a direct relationship between zero-shot NAS scores and the expected performance of the model on unseen datasets within a CL setting. It should be emphasized that the objective here is not to provide a complete solution to the optimization problem formulated earlier, but rather to investigate the potential of adapting NAS-inspired methods to the sequencing problem in continual learning.

\begin{figure}[]
\centering
 {\includegraphics[width=0.45\textwidth]{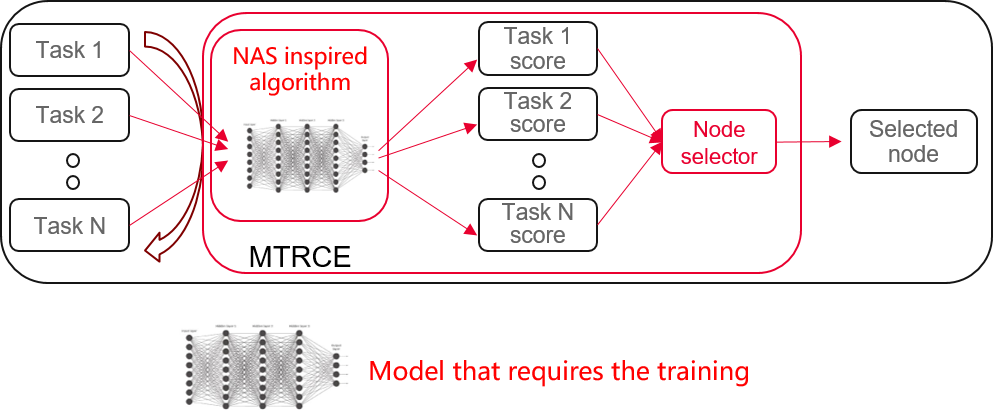}}
 \caption{Framework incorporating NAS-inspired scoring algorithm}
 \label{fig:framework}
\end{figure}

As per the literature, a number of zero-shot methods have been proposed, each with its own advantages and limitations \cite{NAS2}. For the purpose of this work, we curated 11 different zero-shot methods and evaluated their suitability for sequencing in CL. Among these, the method proposed by Mellor et al. in \cite{NWOT}, known as NAS Without Training (NWOT), emerged as the most stable and reliable across diverse experimental conditions. Accordingly, we adopted NWOT as the core component of our sequencing framework. The proposed framework is illustrated in Figure \ref{fig:framework}, which highlights how NWOT is integrated to approximate dataset utility in the CL setting. In the following sections, we report experimental results that demonstrate the feasibility and effectiveness of utilizing concepts from zero-shot NAS methods as part of sequencing strategies in continual learning.

\subsection{NWOT}\label{NWOT}

Mellor et al. \cite{NWOT} introduced a novel approach to NAS that eliminates the need for training candidate networks, thereby drastically reducing computational cost. Their method exploits the overlap of activations between datapoints in untrained networks as a proxy for performance prediction. By analyzing how different inputs produce correlated activations in a randomly initialized architecture, they derived a scoring function that exhibits a strong correlation with the eventual trained accuracy of the network. This insight enables architectures to be evaluated and ranked in seconds on a single GPU, bypassing the traditionally expensive process of training and validating thousands of models in NAS pipelines.

Furthermore, the authors demonstrated that the computed scores evolve consistently as the model undergoes training. Specifically, the score obtained from a randomly initialized model tested against a dataset was lower than the score reported for the same architecture after a few epochs of training. This increase in score provided empirical evidence of a strong correlation between the proposed scoring method and the model’s confidence in its predicted performance. Such findings highlight the robustness of the NWOT algorithm and establish it as a reliable zero-cost proxy for architecture evaluation. Taken together with our experimental results, these observations support the conclusion that NWOT holds significant potential for application in sequencing within CL.

\begin{figure}[t]
   \centering
   \includegraphics[width=0.8\linewidth, height=7cm]{ 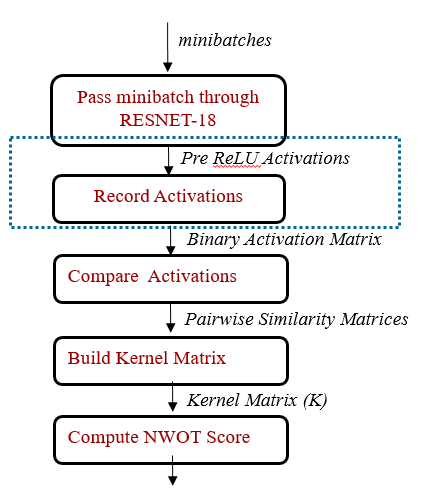}
   \caption{NWOT: Estimating how useful or informative a data minibatch is.}
   \label{nwot}
\end{figure}

The standard NWOT score, as shown in Figure \ref{nwot}, is obtained by: (i) forward-passing a minibatch through the AI model, (ii) recording neuron activations after ReLU layers, (iii) comparing activation patterns across samples to measure similarity, (iv) building a kernel matrix from pairwise similarities, and (v) computing the log-determinant of this kernel as the NWOT score. A higher score indicates greater diversity within the minibatch, and in turn, higher diversity within the associated dataset, which may suggest that it could provide more informative gradients for learning. An important aspect that makes this method appealing is its lightweight and fast computation, as it only requires a single forward pass of a small batch of data.

Accordingly, for this work, and as illustrated in Figure~\ref{fig:framework}, we employ the NWOT scoring algorithm to guide the selection of the next DN in the sequence. Let $R^t = \{r_1^t, r_2^t, \dots, r_N^t\}$ denote the set of NWOT scores computed for the $N$ DNs in the network using the version of the model at time step $t$, where $r_i^t$ represents the score for the $i^{\text{th}}$ DN. To compute $r_i^t$, a small minibatch of data is first collected from the $i^{\text{th}}$ DN. This minibatch is then passed through the model at step $t$, parameterized by weights $\Theta_t$. The neuron activations after the ReLU layers are recorded for each sample in the minibatch, and the similarity between activation patterns is measured to produce the corresponding NWOT score.  

At step $0$, the initial version of the model, denoted by $\Theta_0$, is used to compute the score set $R^0$. These scores are then provided to the \textit{node selector unit}. Since this is the first step and the objective is to maximize performance, the selector deterministically chooses the DN associated with the dataset that yields the highest NWOT score, i.e., $v_h = \max \{R^0\}$.

The model is subsequently trained on the selected dataset, and the weights are updated to $\Theta_1$. A new set of scores, $R^1$, is then computed for all $N$ DNs, including the current DN, and fed back into the selector. In addition, the actual performance metrics of the updated model are reported, such as $M(\Theta_1(S^1))$ and $f_{\Theta_{1}(S^1)}(D_{\text{test}}^{v_h,\text{sub}})$.  

At this stage, the selector has access to multiple sources of information: the initial scores $R^0$, the updated scores $R^1$, and the observed performance metrics. This allows for better decision-making strategies. For example, the selector may analyze the rate of change of scores across DNs to anticipate potential CF. If the scores for a particular DN are observed to decline rapidly as training progresses, the selector may prioritize that DN next to mitigate CF effects. Alternatively, the selector may adopt a \textit{rotational strategy}, sequentially selecting DNs in descending order of their scores (e.g., round 1: highest score, round 2: second highest score, \dots, round $n$: $n^{th}$ highest score, then repeat).  Algorithm \ref{alg:NWOTalgo} describes the process in a step-by-step manner.

\begin{algorithm}[t]
\caption{NWOT-Guided Node Selection in CL}
\label{alg:NWOTalgo}
\begin{algorithmic}[1]
\State \textbf{Input:} Model parameters $\Theta_0$, DNs $\{v_1,\dots,v_N\}$, horizon $T$
\State \textbf{Output:} Updated parameters $\Theta_T$ after sequencing
\For{$t = 0$ \textbf{to} $T-1$}
    \State \textbf{Compute NWOT scores for all DNs}
    \For{\textbf{each} DN $v_i$}
        \State Sample minibatch $d_i^t$ from DN $v_i$
        \State Forward pass with $\Theta_t$ on $d_i^t$
        \State Record ReLU activations for all samples
        \State Compute similarity of activation patterns
        \State Derive NWOT score $r_i^t$
    \EndFor
    \State Construct $R^t = \{r_1^t, r_2^t, \dots, r_N^t\}$
    \State \textbf{Selector decision}
    \If{$t=0$}
        \State $v_h \gets \arg\max\{R^0\}$
    \Else
        \State Analyze $R^{t-1}$, $R^t$, and metrics $M(\Theta_t)$
        \State Optionally compute $\Delta r_i \gets r_i^t - r_i^{t-1}$
        \State Apply strategy (highest score, CF balancing, rotation)
        \State Select next DN $v_h$
    \EndIf
    \State \textbf{Model update}
    \State Train on dataset of $v_h$
    \State Update parameters $\Theta_{t+1}$
    \State Report $M(\Theta_{t+1})$ and $f_{\Theta_{t+1}}(D_{\text{test}}^{v_h,\text{sub}})$
\EndFor
\end{algorithmic}
\end{algorithm}

Ultimately, the specific selection logic is flexible and can be tailored to the application. The overarching objective is to leverage NWOT scores, in conjunction with performance metrics, as inputs to the selector to make informed and adaptive decisions about the next DN in the sequence. This framework provides a principled mechanism for balancing performance maximization with CF mitigation in CL. It should be noted, however, and as will be shown in the results section, that the original NWOT scoring method and the above proposed framework work well in the case of independent and identically distributed (I.I.D.) data or tasks. In the case of non-I.I.D. data, NWOT in its original form produces inconsistent values. As such, in the next section, we detail a modification that can be applied to NWOT to make it more useful for the case of non-I.I.D tasks.

\subsection{NWOT modified} \label{scaling}

\begin{figure}[t]
   \centering
   \includegraphics[width=\linewidth]{ 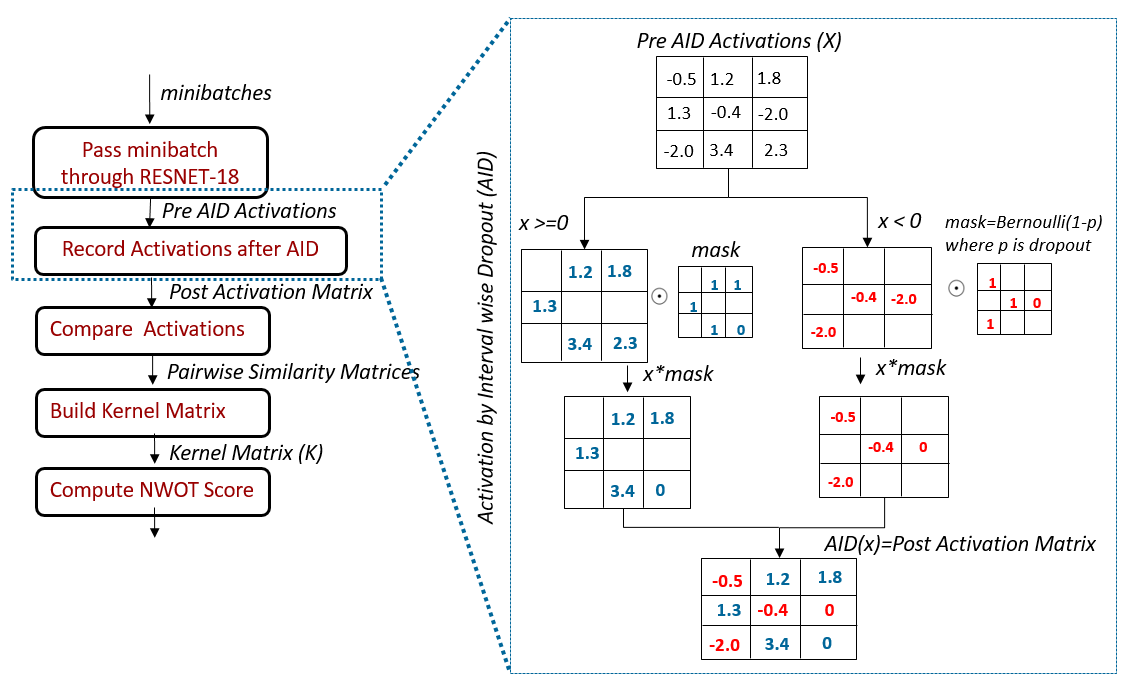}
   \caption{AID-based NWOT Scoring: ReLU activations replaced by AID}
   \label{nwot-AID}
\end{figure}

In real-world distributed environments, data and tasks are rarely I.I.D. Instead, each DN typically exhibits unique statistical properties, leading to heterogeneous feature spaces, class imbalances, and varying domain shifts. This non-I.I.D nature of data poses significant challenges for traditional training and model selection strategies, as models often overfit to dominant domains while underperforming on underrepresented ones. To address this limitation, we extend the NWOT scoring method described in Section~\ref{NWOT} to explicitly account for non-I.I.D settings.

As previously discussed, the standard NWOT formulation produces scores that are proportional to the diversity observed within a minibatch of data. A higher score generally indicates greater diversity among samples, which in turn suggests that the associated training dataset may provide more informative gradients for learning. However, in non-I.I.D scenarios, this scoring formulation exhibits poor robustness. For example, datasets from domains such as sketch or quickdraw in the DomainNet benchmark naturally yield low-variance activations after ReLU, leading NWOT to assign them misleadingly low scores compared to domains such as infograph or real. Directly comparing scores across these inherently different domains may result in erroneous conclusions.

As shown in Table \ref{comparison}, the NWOT scores computed using a randomly initialized ResNet18 for the domains real, infograph, sketch, and quickdraw suggest that the infograph and real domains (with lower scores) are less informative than the sketch and quickdraw domains (with higher scores). However, experimental evidence indicates otherwise. For example, in an experiment with a ResNet18 model trained for 15 epochs on the quickdraw domain, the model achieved approximately 75\% accuracy on quickdraw but only an average global performance of 18.5\%. In contrast, when trained on the real domain for the same duration, the model reached about 42\% accuracy on real but achieved a higher average global performance of 23\%. This demonstrates that knowledge gained from the real domain was more transferable than that from quickdraw, as training on real enabled the model to learn features such as color, which improved recognition across other domains and led to superior global performance.

\begin{table}[]
\caption{misleading results for non-I.I.D scenario}
\label{comparison}
\begin{tabular}{|l|l|l|l|l|}
\hline
\textbf{Domain}     & \textbf{Real} & \textbf{Infograph} & \textbf{Sketch} & \textbf{Quickdraw} \\ \hline
\textbf{NWOT score} & 1738.6        & 1729.7             & 1788.3          & 1803.4             \\ \hline
\textbf{\begin{tabular}[c]{@{}l@{}}Per-domain accuracy\\ performance \\ (15 epochs)\end{tabular}} & 41.96\% & 36.7\% & 58.9\%  & 75.1\%  \\ \hline
\textbf{\begin{tabular}[c]{@{}l@{}}Global accuracy \\ performance\\ (15 epochs)\end{tabular}}     & 23.1\%  & 20.9\% & 15.75\% & 18.46\% \\ \hline
\end{tabular}
\end{table}

This discrepancy in NWOT score scaling under non-I.I.D conditions can lead to repeated over-selection of certain domains and neglect of others, ultimately biasing the training process and increasing catastrophic forgetting. To mitigate this issue, modifications to NWOT are required to regularize activations and ensure that the score reflects meaningful diversity rather than artifacts of domain-specific variance.

Therefore, we modify NWOT by introducing activation interval dropout (AID) \cite{AID} at the activation stage, as shown in Fig. \ref{nwot-AID}. Instead of directly using raw ReLU output, activations are perturbed by interval-wise dropout, where each predefined activation interval is assigned its own dropout probability. Let the pre-activation matrix be defined by $X \in \mathbb{R}^{n \times d}$ with entries $X_{ij}$. AID defines disjoint intervals $\{I_1, I_2, \dots, I_k\}$ with the corresponding dropout probabilities $\{p_1, p_2, \dots, p_k\}$. For each element $X_{ij}$ within each interval, a binary mask $m_{ij}$ is generated as  
\begin{equation}
    m_{ij} \sim \text{Bernoulli}(1 - p_\ell), \quad \text{if } X_{ij} \in I_\ell
\end{equation}
where $p_\ell$ is the corresponding probability of interval from which $x_{ij}$ belongs. The perturbed activation is then computed as:
\begin{equation}
    \tilde{X}_{ij} = X_{ij} \cdot m_{ij} 
\end{equation}

In our non-I.I.D setting, we use two intervals as $I_1 = (-\infty, 0), \quad I_2 = [0, \infty)$ with dropout probabilities $p_1$ for negative values and $p_2$ for positive values. The resulting AID-regularized activation matrix $\tilde{X}$ replaces the raw ReLU outputs in the NWOT computation. The subsequent steps, pairwise similarity computation, kernel construction, and log-determinant calculation, remain unchanged. This structured stochastic masking prevents the kernel computation from being biased toward sharp or domain-specific activation patterns. 

Formally, the activation under AID can be expressed as
\begin{equation}
    \tilde{\sigma}(X)=
    \begin{cases}
    x_{ij} \cdot m \quad x_{ij} \geq 0 \quad m \sim  \text{Bernoulli}(p_\ell) \\
    x_{ij} \cdot m \quad x_{ij} < 0 \quad m \sim 
    \text{Bernoulli}(1-p_\ell)
    \end{cases}
\end{equation}
where $x_{ij}$ is the pre-activation input, $p_\ell$ is the interval-specific dropout probability and $m$ is the binary mask. 


\section{results}
In this section, the experimental results are presented. All the results are based on ResNet18 model and data from the DomainNet dataset.
\subsection{I.I.D-NWOT}
\begin{figure}[]
   \centering
   \includegraphics[width=\linewidth]{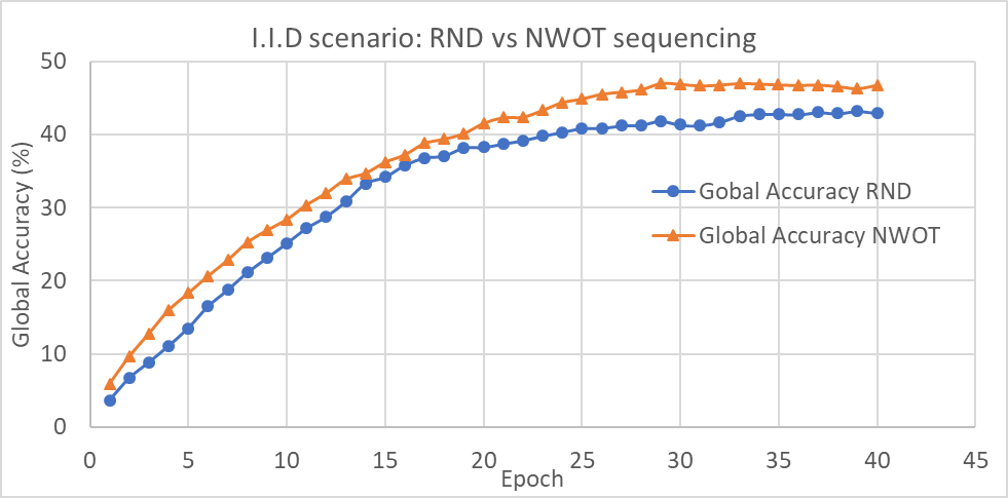}
   \caption{Sequencing using NWOT in the case of I.I.D}
   \label{NWOT-iid-results}
\end{figure}

In the I.I.D. scenario, data from all domains within the DomainNet dataset is combined and split into 70\% training samples (approximately 400k) and 30\% testing samples (approximately 175k). The training set is then randomly partitioned into six sub-training datasets, simulating a KSN with six DNs. Because the samples are randomly divided, this setup constitutes a semi-I.I.D. scenario, with each DN receiving a nearly identical distribution of data. A ResNet18 model is employed to sequentially learn from the six DNs. Following the framework described in Section \ref{sequencing_framework}, a minibatch is collected from each DN, and the NWOT score is computed using the original formulation from \cite{NWOT}. This process is repeated over 40 training epochs. For comparison, a baseline was established by generating a random sequence of 40 steps, which the model followed to learn the six datasets. The results, shown in Figure \ref{NWOT-iid-results}, represent the average of five repetitions of the experiment. For each repetition, the dataset division was performed from scratch, and a new random sequence was generated. The model initialization was also changed each time, though it remained randomly initialized. Importantly, within each experiment, the same initial model was used for both the random sequence and the NWOT sequence to ensure a fair comparison.

From the results shown in Figure \ref{NWOT-iid-results}, it can be observed that CF did not occur in this scenario. This outcome is expected, given the semi-I.I.D. nature of the data, where samples from each domain and every class are present at each DN. Nevertheless, the results clearly indicate that random sequencing led to approximately 7-9\% lower overall global accuracy across the 40 training epochs.

\subsection{NON-I.I.D-NWOT}
\begin{figure}[t]
   \centering
   \includegraphics[width=\linewidth]{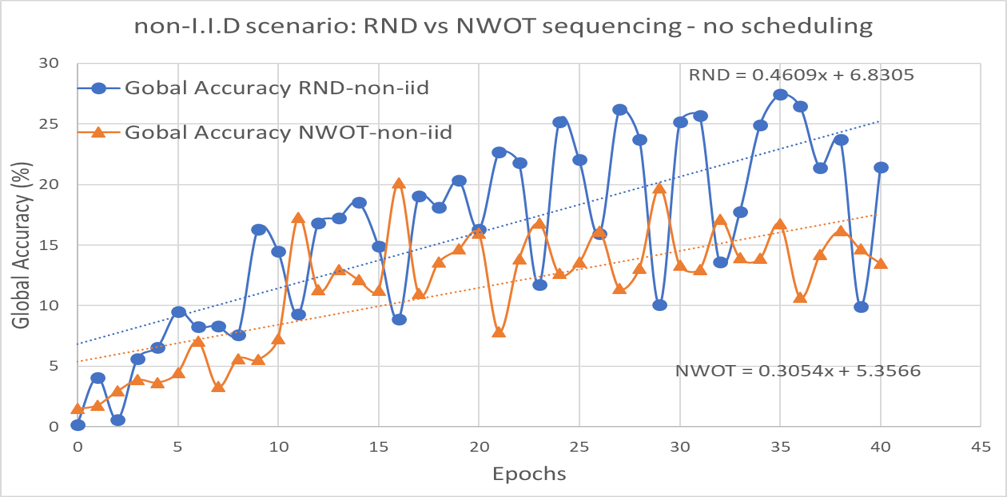}
   \caption{Sequencing using NWOT in the case of non-I.I.D without proper scheduling}
   \label{NWOT-non-iid-1}
\end{figure}

Similarly, a non-I.I.D. scenario was established using data from the DomainNet dataset. In this setup, each of the six domains within DomainNet was treated as a separate task and assigned to a DN in the KSN, effectively creating a domain-based CL problem. A randomly initialized ResNet18 model was then employed to sequentially learn the six domains. As in the I.I.D. scenario, the baseline consisted of random sequencing: a random sequence of 40 steps was generated, and the model followed that sequence while learning the different domains. The experiment was repeated five times, and the average global accuracy performance is presented in Figure \ref{NWOT-non-iid-1}. For each repetition, a new random sequence was generated, and a newly initialized model was used. The same initial model was used for both the random sequence and the NWOT-based sequencing. It should be clarified that the global accuracy shown in the figure represents the average performance across all domains. Each data point in the graph was computed after the model had trained on the selected node at that epoch.

It is clear from the results shown in the figure that CF was severe for both the random sequencing and the NWOT-based sequencing, as evidenced by the highly fluctuating performance. It is also clear that the global accuracy is effectively lower in this case compared to the case of I.I.D for both sequencing methods, which is an expected outcome as data samples from the different domains are not present everywhere. An important behaviour to report is that the random sequences were generated such that all six domains will be visited at least once during the 40 epochs of training. When NWOT was used, the sequences generated seemed to have always ignored visiting inforgraph and painting domains, have sporadically visited the real and quickdraw domains, but has almost equally visited the sketch and clipart domains. This is also an expected behaviour due to the NWOT scaling problem described in Section \ref{scaling}. This behaviour resulted in NWOT sequencing being inferior to random sequencing, as shown in the figure.

\begin{figure}[]
   \centering
   \includegraphics[width=\linewidth]{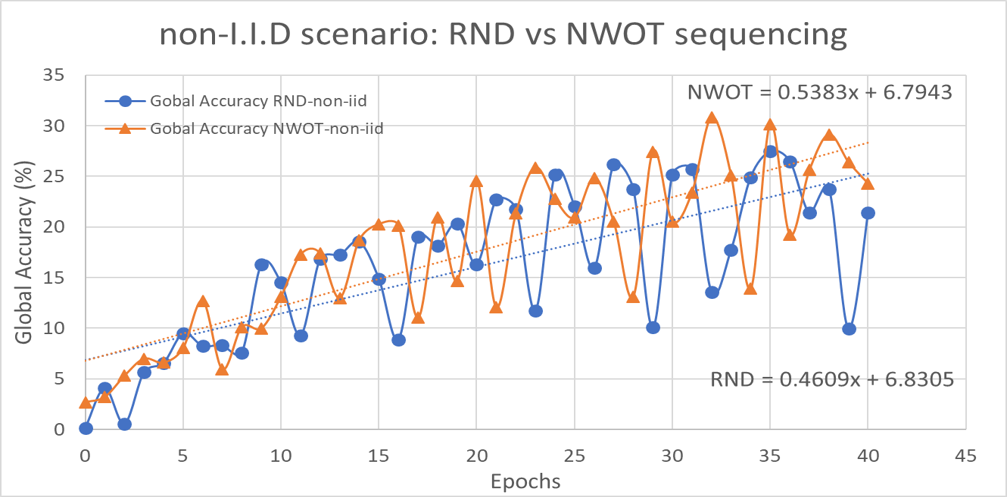}
   \caption{Sequencing using NWOT in the case of non-I.I.D with scheduling}
   \label{NWOT-non-iid}
\end{figure}

To counteract this issue, a scheduler was designed as per Algorithm 1 described in Section  \ref{NWOT}. Essentially, we force the model to visit all nodes by observing the frequency of visits to each node as well as the assocaited CF behaviour. In doing so, the above experiment was repeated, and the results are as shown in Figure \ref{NWOT-non-iid}.

As shown in the figure, while CF was still severe, by implementing a scheduler, NWOT sequencing was able to perform slightly better than random sequencing. By observing the trendline for the graphs, it is clear that the NWOT sequencing resulted in an overall better trend with a faster convergence rate. Yet, it is clear that sequencing did not eliminate CF, but rather reduced its impact. 

\begin{figure}[]
   \centering
   \includegraphics[width=\linewidth]{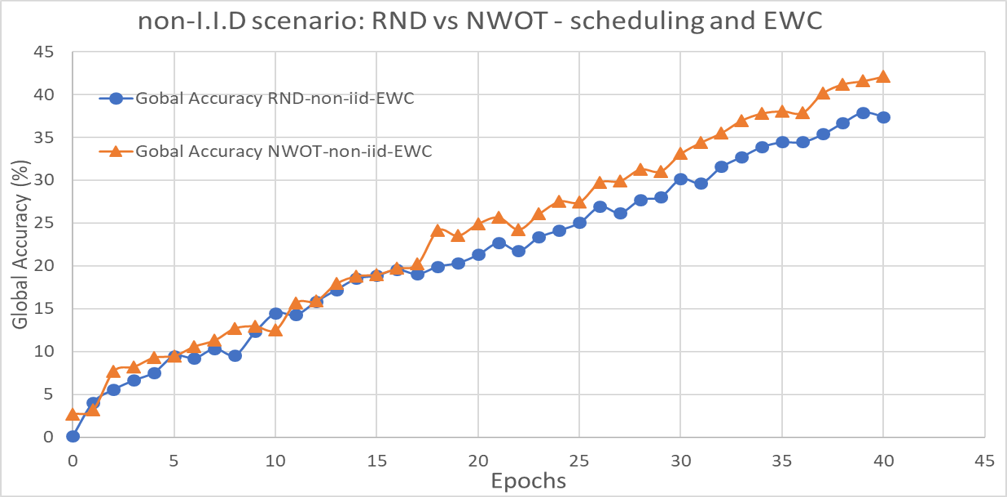}
   \caption{Sequencing using NWOT in the case of non-I.I.D with scheduling and EWC}
   \label{EWC}
\end{figure}

Motivated by this result, in the following, the NWOT sequencing is coupled with a traditional CL method. To be exact, we couple the NWOT with elastic weight consolidation (EWC) from \cite{49-2} to study the impact of sequencing in combination with CL methods. The same non-I.I.D setup is used along with the implemented scheduler. The main difference is that, before training the model on the selected next DN, EWC is applied to the training step. This is done for both the random and NWOT-based sequencing. The results are shown in Figure \ref{EWC}. The results shown are the average of five repetitions of the experiment as previously described.

It is clear from the results shown in the figure that when applying effective sequencing along with a traditional CL-based method such as EWC, the impact of CF is not only mitigated, but the overall accuracy performance is improved as evidenced by the improved accuracy of the NWOT-based sequencing.

\subsection{NON-I.I.D-NWOT-AID}
\begin{figure*}[htbp]
    \centering
    
    \begin{subfigure}{0.22\textwidth}
        \includegraphics[width=\linewidth,  height=4cm]{ 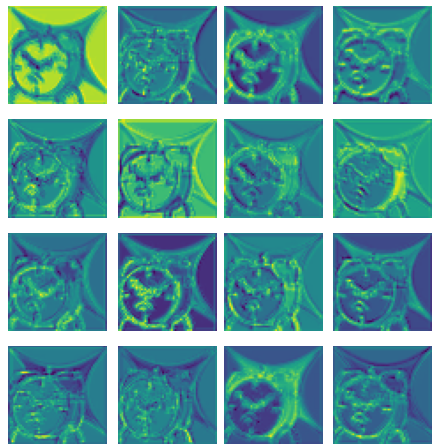}
        \caption{Clipart Activation Map}
        \label{relu:sub-a}
    \end{subfigure}\hfill
    \begin{subfigure}{0.22\textwidth}
        \includegraphics[width=\linewidth,  height=4cm]{ 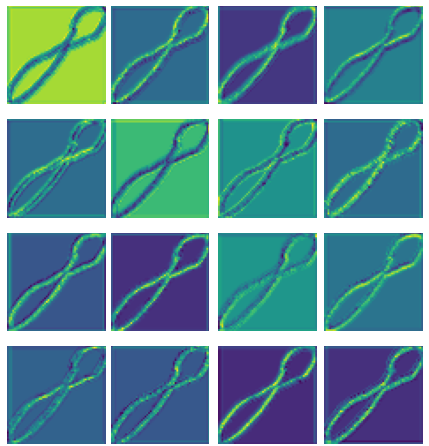}
        \caption{Quickdraw Activation Map}
        \label{relu:sub-b}
    \end{subfigure}\hfill
    \begin{subfigure}{0.22\textwidth}
        \includegraphics[width=\linewidth,  height=4cm]{ 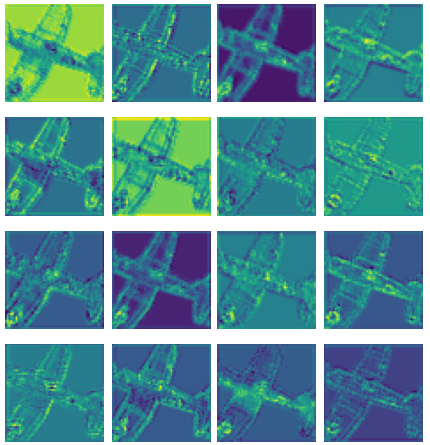}
        \caption{Sketch Activation Map}
        \label{relu:sub-c}
    \end{subfigure}
    \hfill
    \begin{subfigure}{0.22\textwidth}
        \includegraphics[width=\linewidth,  height=4cm]{ 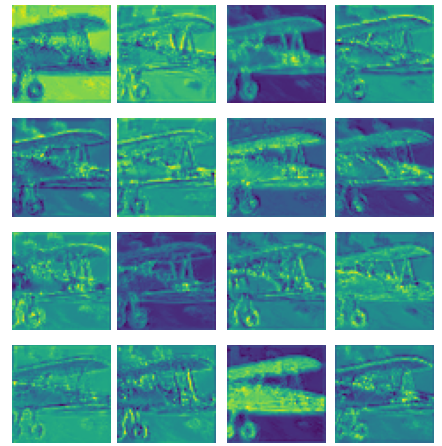}
        \caption{Painting Activation Map}
        \label{relu:sub-d}
    \end{subfigure}

    \begin{subfigure}{0.22\textwidth}
        \includegraphics[width=\linewidth,  height=4cm]{ 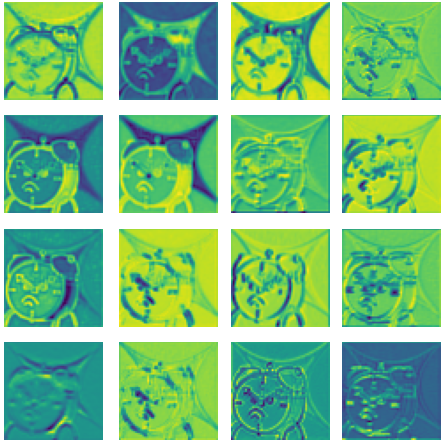}
        \caption{Clipart Activation Map}
        \label{aid:sub-a}
    \end{subfigure}
    \hfill
    \begin{subfigure}{0.22\textwidth}
        \includegraphics[width=\linewidth, height=4cm]{ 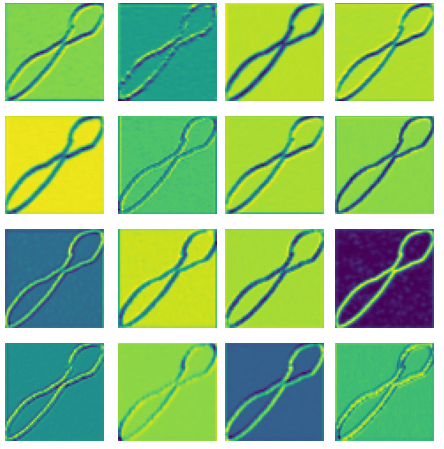}
        \caption{Quickdraw Activation Map}
        \label{aid:sub-b}
    \end{subfigure}
    \hfill
    \begin{subfigure}{0.22\textwidth}
        \includegraphics[width=\linewidth ,  height=4cm]{ 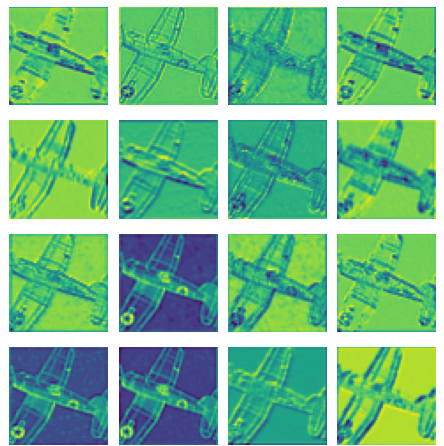}
        \caption{Sketch Activation Map}
        \label{aid:sub-c}
    \end{subfigure}
    \hfill
    \begin{subfigure}{0.22\textwidth}
        \includegraphics[width=\linewidth,  height=4cm]{ 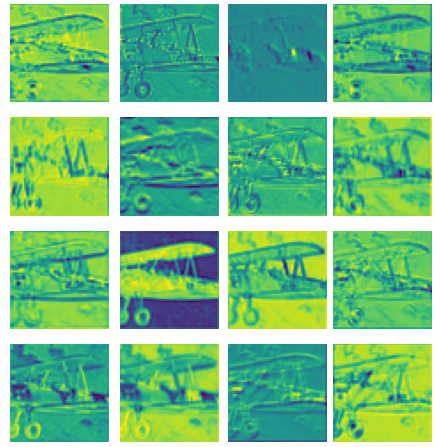}
        \caption{Painting Activation Map}
        \label{aid:sub-d}
    \end{subfigure}

    \caption{\texttt{ReLU} vs. \texttt{Activation Interval Dropout (AID)} based Post Activation Maps for DomainNet\cite{domainnet}}
    \label{Activations}
\end{figure*}

To evaluate the outcome of AID-based activations for non-I.I.D dataset, we plot the activation maps for four different domains as clipart, sketch, quickdraw and painting using ResNet-18 as shown in Fig. \ref{Activations}.
It can be seen that activation maps are presented for both \texttt{ReLU} and \texttt{AID}. The \texttt{ReLU} preserves only positive responses while discarding negative values, resulting in concentration of activations in limited regions. This indicates reduced diversity in the feature space, which leads \texttt{NWOT} to favor domains disproportionately during selection. Consequently, the ReLU-driven \texttt{NWOT} tends to overfit to sparse or low-complexity domains, undermining generalization across heterogeneous DNs.

In contrast, the \texttt{AID} activations exhibit more heterogeneous and distributed responses across domains, with reduced dominance of a few saturated neurons and greater variance in activation magnitudes. By introducing interval-wise dropout \texttt{AID} at the activation stage, \texttt{AID} perturbs neurons differently depending on their activation interval, thereby preventing the same set of high-activation neurons from repeatedly dominating the similarity structure. This leads to 
higher log-determinant scores for underrepresented domains, ensuring that \texttt{NWOT} selection is less biased and more balanced across heterogeneous DNs.

\begin{figure}[t]
   \centering
   \includegraphics[width=\linewidth]{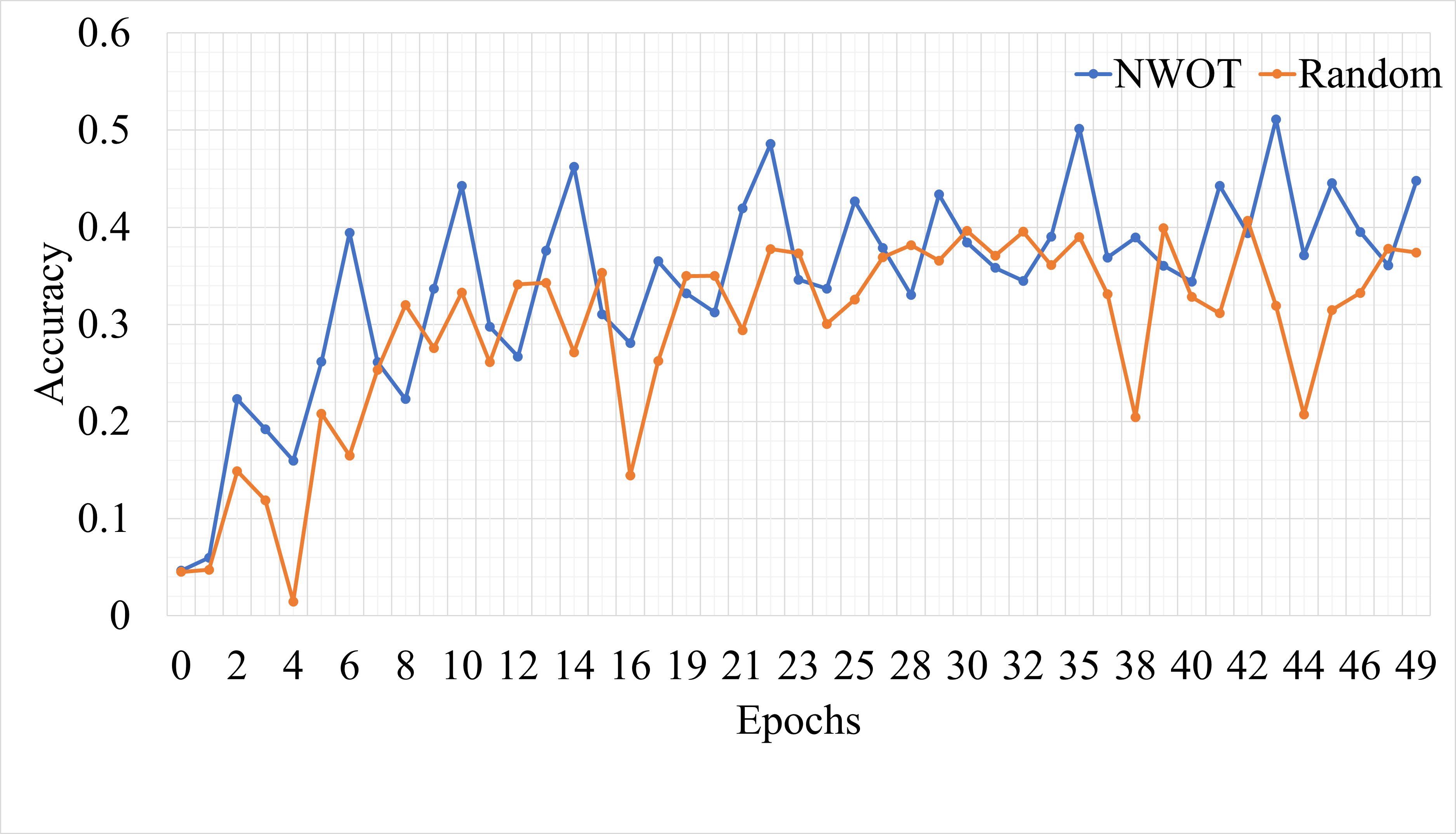}
   \caption{Evaluation Accuracy of ResNet-18 using NWOT-AID and Random selection of Nodes}
   \label{nwot-random}
\end{figure}

The same non-I.I.D setup is used for testing the NWOT-AID approach. No EWC or scheduler were used for this part. Figure \ref{nwot-random} presents the evaluation accuracy of ResNet-18 under non-I.I.D sequential training when domains are selected either randomly or through the proposed \texttt{NWOT-AID} scoring mechanism. The random baseline shows gradual improvement, reaching an accuracy plateau around 35–38\%, with highly unstable fluctuations. This instability reflects the inefficiency of sampling domains without considering diversity, which often results in redundant updates from low complexity domains, such as sketch. On the contrary, the \texttt{NWOT-AID} selection method achieves higher accuracy of around 50\%, and the trajectory to achieve this accuracy remains stable than the random, highlighting that \texttt{NWOT-AID} mitigates the instability associated with the non-I.I.D sequential learning. Furthermore, \texttt{NWOT-AID} capture feature diversity more effectively, allowing the selection to prioritize informative domains. Moreover, it reduces bias towards sparse domains and accelerates convergence. It can be concluded from this result that the AID was able to resolve the scaling problem of the original NWOT described in Section \ref{NWOT}.

\section{Conclusions and Future work}

In this paper, we addressed catastrophic forgetting in continual learning by proposing a lightweight method inspired by neural architecture search, called NWOT. NWOT helps determine the optimal next node in a sequence to maximize global performance while minimizing forgetting. We formulated the problem as a stochastic optimization and highlighted the difficulty of predicting performance and forgetting without prior training. Our experiments showed that sequencing can be guided using a zero-shot NAS scoring method, which improves performance in I.I.D. scenarios but struggles in non-I.I.D. settings. To address these shortcomings, we introduced two enhancements: (i) a scheduler that leverages historical values and CF behavior for node selection, and (ii) activation interval dropout (AID), which diversifies neuron responses and reduces bias by improving scores for underrepresented domains. We further demonstrated that coupling NWOT sequencing with traditional CL methods such as EWC enhances their performance. Although the study was limited to a specific dataset and model type, the results suggest clear advantages to sequencing with predictive methods like zero-shot NAS. Future work will explore alternative prediction strategies, including Fisher-value-based approaches. Additionally, integrating these approaches into the AI network proposed in \cite{hesham} could be an important next step towards building a coherent AI ecosystem.

\bibliographystyle{IEEEtran}
\bibliography{References.bib}

@inproceedings{16,
  title={icarl: Incremental classifier and representation learning},
  author={Rebuffi, Sylvestre-Alvise and Kolesnikov, Alexander and Sperl, Georg and Lampert, Christoph H},
  booktitle={Proceedings of the IEEE conference on Computer Vision and Pattern Recognition},
  pages={2001--2010},
  year={2017}
}

@ARTICLE{hesham,
  author={Moussa, Hesham G. and Akhavain, Arashmid and Maryam Hosseini, S. and McCormick, Bill},
  journal={IEEE Network}, 
  title={Distributed Learning and Inference Systems: A Networking Perspective}, 
  year={2025},
  volume={},
  number={},
  pages={1-1},
  doi={10.1109/MNET.2025.3573940}}

@article{49,
  title={Experience replay for continual learning},
  author={Rolnick, David and Ahuja, Arun and Schwarz, Jonathan and Lillicrap, Timothy and Wayne, Gregory},
  journal={Advances in neural information processing systems},
  volume={32},
  year={2019}
}

@inproceedings{50,
  title={Selective experience replay for lifelong learning},
  author={Isele, David and Cosgun, Akansel},
  booktitle={Proceedings of the AAAI Conference on Artificial Intelligence},
  volume={32},
  number={1},
  year={2018}
}

@inproceedings{51,
  title={Continual learning with tiny episodic memories},
  author={Chaudhry, Arslan and Rohrbach, Marcus and Elhoseiny, Mohamed and Ajanthan, Thalaiyasingam and Dokania, P and Torr, P and Ranzato, M},
  booktitle={Workshop on Multi-Task and Lifelong Reinforcement Learning},
  year={2019}
}

@article{55,
  title={Gradient episodic memory for continual learning},
  author={Lopez-Paz, David and Ranzato, Marc'Aurelio},
  journal={Advances in neural information processing systems},
  volume={30},
  year={2017}
}

@misc{bell,
      title={The Effect of Task Ordering in Continual Learning}, 
      author={Samuel J. Bell and Neil D. Lawrence},
      year={2022},
      eprint={2205.13323},
      archivePrefix={arXiv},
      primaryClass={cs.LG},
      url={https://arxiv.org/abs/2205.13323}, 
}

@article{6,
  title={Efficient lifelong learning with a-gem},
  author={Chaudhry, Arslan and Ranzato, Marc'Aurelio and Rohrbach, Marcus and Elhoseiny, Mohamed},
  journal={arXiv preprint arXiv:1812.00420},
  year={2018}
}

@article{118,
  title={Learning without forgetting},
  author={Li, Zhizhong and Hoiem, Derek},
  journal={IEEE Transactions on pattern analysis and machine intelligence},
  volume={40},
  number={12},
  pages={2935--2947},
  year={2017},
  publisher={IEEE}
}

@inproceedings{119,
  title={Learning without memorizing},
  author={Dhar, Prithviraj and Singh, Rajat Vikram and Peng, Kuan-Chuan and Wu, Ziyan and Chellappa, Rama},
  booktitle={Proceedings of the IEEE/CVF conference on computer vision and pattern recognition},
  pages={5138--5146},
  year={2019}
}

@misc{108,
      title={Rotate your Networks: Better Weight Consolidation and Less Catastrophic Forgetting}, 
      author={Xialei Liu and Marc Masana and Luis Herranz and Joost Van de Weijer and Antonio M. Lopez and Andrew D. Bagdanov},
      year={2018},
      eprint={1802.02950},
      archivePrefix={arXiv},
      primaryClass={cs.CV},
      url={https://arxiv.org/abs/1802.02950}, 
}

@inproceedings{109,
  title={Continual learning with extended kronecker-factored approximate curvature},
  author={Lee, Janghyeon and Hong, Hyeong Gwon and Joo, Donggyu and Kim, Junmo},
  booktitle={Proceedings of the IEEE/CVF Conference on Computer Vision and Pattern Recognition},
  pages={9001--9010},
  year={2020}
}

@article{NAS1,
  title={AZ-NAS: Assembling Zero-Cost Proxies for Network Architecture Search},
  author={Junghyup Lee and Bumsub Ham},
  journal={2024 IEEE/CVF Conference on Computer Vision and Pattern Recognition (CVPR)},
  year={2024},
  pages={5893-5903},
  url={https://api.semanticscholar.org/CorpusID:268732878}
}

@article{NAS2,
  title={NAS-Bench-Suite-Zero: Accelerating Research on Zero Cost Proxies},
  author={Arjun Krishnakumar and Colin White and Arber Zela and Renbo Tu and Mahmoud Safari and Frank Hutter},
  journal={ArXiv},
  year={2022},
  volume={abs/2210.03230},
  url={https://api.semanticscholar.org/CorpusID:252762140}
}

@ARTICLE{NAS-survery,
  author={Li, Guihong and Hoang, Duc and Bhardwaj, Kartikeya and Lin, Ming and Wang, Zhangyang and Marculescu, Radu},
  journal={IEEE Transactions on Pattern Analysis and Machine Intelligence}, 
  title={Zero-Shot Neural Architecture Search: Challenges, Solutions, and Opportunities}, 
  year={2024},
  volume={46},
  number={12},
  pages={7618-7635},
  doi={10.1109/TPAMI.2024.3395423}}

@inproceedings{NWOT,
  title={Neural architecture search without training},
  author={Mellor, Joe and Turner, Jack and Storkey, Amos and Crowley, Elliot J},
  booktitle={International conference on machine learning},
  pages={7588--7598},
  year={2021},
  organization={PMLR}
}

@article{CuL3,
  title={Sequence Transferability and Task Order Selection in Continual Learning},
  author={Nguyen, Thinh and Nguyen, Cuong N and Pham, Quang and Nguyen, Binh T and Ramasamy, Savitha and Li, Xiaoli and Nguyen, Cuong V},
  journal={arXiv preprint arXiv:2502.06544},
  year={2025}
}

@inproceedings{CuL2,
  title={Overcoming catastrophic forgetting in massively multilingual continual learning},
  author={Winata, Genta Indra and Xie, Lingjue and Radhakrishnan, Karthik and Wu, Shijie and Jin, Xisen and Cheng, Pengxiang and Kulkarni, Mayank and Preo{\c{t}}iuc-Pietro, Daniel},
  booktitle={Findings of the Association for Computational Linguistics: ACL 2023},
  pages={768--777},
  year={2023}
}

@ARTICLE{CuL1,
  author={Wang, Xin and Chen, Yudong and Zhu, Wenwu},
  journal={IEEE Transactions on Pattern Analysis and Machine Intelligence}, 
  title={A Survey on Curriculum Learning}, 
  year={2022},
  volume={44},
  number={9},
  pages={4555-4576},
  doi={10.1109/TPAMI.2021.3069908}}

@ARTICLE{LLmCF2,
  author={Luo, Yun and Yang, Zhen and Meng, Fandong and Li, Yafu and Zhou, Jie and Zhang, Yue},
  journal={IEEE Transactions on Audio, Speech and Language Processing}, 
  title={An Empirical Study of Catastrophic Forgetting in Large Language Models During Continual Fine-Tuning}, 
  year={2025},
  volume={33},
  number={},
  pages={3776-3786},
  doi={10.1109/TASLPRO.2025.3606231}}

@article{llmCF1,
  title={Mitigating catastrophic forgetting in large language models with self-synthesized rehearsal},
  author={Huang, Jianheng and Cui, Leyang and Wang, Ante and Yang, Chengyi and Liao, Xinting and Song, Linfeng and Yao, Junfeng and Su, Jinsong},
  journal={arXiv preprint arXiv:2403.01244},
  year={2024}
}

@misc{surveycl,
      title={A Comprehensive Survey of Continual Learning: Theory, Method and Application}, 
      author={Liyuan Wang and Xingxing Zhang and Hang Su and Jun Zhu},
      year={2024},
      eprint={2302.00487},
      archivePrefix={arXiv},
      primaryClass={cs.LG},
      url={https://arxiv.org/abs/2302.00487}, 
}

@article{49-2,
  title={Overcoming catastrophic forgetting in neural networks},
  author={Kirkpatrick, James and Pascanu, Razvan and Rabinowitz, Neil and Veness, Joel and Desjardins, Guillaume and Rusu, Andrei A and Milan, Kieran and Quan, John and Ramalho, Tiago and Grabska-Barwinska, Agnieszka and others},
  journal={Proceedings of the national academy of sciences},
  volume={114},
  number={13},
  pages={3521--3526},
  year={2017},
  publisher={National Acad Sciences}
}

@inproceedings{67,
  title={Orthogonal gradient descent for continual learning},
  author={Farajtabar, Mehrdad and Azizan, Navid and Mott, Alex and Li, Ang},
  booktitle={International Conference on Artificial Intelligence and Statistics},
  pages={3762--3773},
  year={2020},
  organization={PMLR}
}

@article{86,
  title={Linear mode connectivity in multitask and continual learning},
  author={Mirzadeh, Seyed Iman and Farajtabar, Mehrdad and Gorur, Dilan and Pascanu, Razvan and Ghasemzadeh, Hassan},
  journal={arXiv preprint arXiv:2010.04495},
  year={2020}
}

@misc{113,
      title={Towards Better Plasticity-Stability Trade-off in Incremental Learning: A Simple Linear Connector}, 
      author={Guoliang Lin and Hanlu Chu and Hanjiang Lai},
      year={2022},
      eprint={2110.07905},
      archivePrefix={arXiv},
      primaryClass={cs.LG},
      url={https://arxiv.org/abs/2110.07905}, 
}

@misc{221,
      title={Piggyback: Adapting a Single Network to Multiple Tasks by Learning to Mask Weights}, 
      author={Arun Mallya and Dillon Davis and Svetlana Lazebnik},
      year={2018},
      eprint={1801.06519},
      archivePrefix={arXiv},
      primaryClass={cs.CV},
      url={https://arxiv.org/abs/1801.06519}, 
}

@article{227,
  title={Lifelong learning with dynamically expandable networks},
  author={Yoon, Jaehong and Yang, Eunho and Lee, Jeongtae and Hwang, Sung Ju},
  journal={arXiv preprint arXiv:1708.01547},
  year={2017}
}

@misc{228,
      title={Compacting, Picking and Growing for Unforgetting Continual Learning}, 
      author={Steven C. Y. Hung and Cheng-Hao Tu and Cheng-En Wu and Chien-Hung Chen and Yi-Ming Chan and Chu-Song Chen},
      year={2019},
      eprint={1910.06562},
      archivePrefix={arXiv},
      primaryClass={cs.LG},
      url={https://arxiv.org/abs/1910.06562}, 
}

@article{238,
  title={Scalable and order-robust continual learning with additive parameter decomposition},
  author={Yoon, Jaehong and Kim, Saehoon and Yang, Eunho and Hwang, Sung Ju},
  journal={arXiv preprint arXiv:1902.09432},
  year={2019}
}

@article{244,
  title={Random path selection for continual learning},
  author={Rajasegaran, Jathushan and Hayat, Munawar and Khan, Salman H and Khan, Fahad Shahbaz and Shao, Ling},
  journal={Advances in neural information processing systems},
  volume={32},
  year={2019}
}

@misc{yolo,
      title={YOLOv11: An Overview of the Key Architectural Enhancements}, 
      author={Rahima Khanam and Muhammad Hussain},
      year={2024},
      eprint={2410.17725},
      archivePrefix={arXiv},
      primaryClass={cs.CV},
      url={https://arxiv.org/abs/2410.17725}, 
}

@article{gpt,
   title={Summary of ChatGPT-Related research and perspective towards the future of large language models},
   volume={1},
   ISSN={2950-1628},
   url={http://dx.doi.org/10.1016/j.metrad.2023.100017},
   DOI={10.1016/j.metrad.2023.100017},
   number={2},
   journal={Meta-Radiology},
   publisher={Elsevier BV},
   author={Liu, Yiheng and Han, Tianle and Ma, Siyuan and Zhang, Jiayue and Yang, Yuanyuan and Tian, Jiaming and He, Hao and Li, Antong and He, Mengshen and Liu, Zhengliang and Wu, Zihao and Zhao, Lin and Zhu, Dajiang and Li, Xiang and Qiang, Ning and Shen, Dingang and Liu, Tianming and Ge, Bao},
   year={2023},
   month=sep, pages={100017} }

@article{alphazero,
  title={Acquisition of chess knowledge in alphazero},
  author={McGrath, Thomas and Kapishnikov, Andrei and Toma{\v{s}}ev, Nenad and Pearce, Adam and Wattenberg, Martin and Hassabis, Demis and Kim, Been and Paquet, Ulrich and Kramnik, Vladimir},
  journal={Proceedings of the National Academy of Sciences},
  volume={119},
  number={47},
  pages={e2206625119},
  year={2022},
  publisher={National Acad Sciences}
}

@inproceedings{domainnet,
  title={Moment matching for multi-source domain adaptation},
  author={Peng, Xingchao and Bai, Qinxun and Xia, Xide and Huang, Zijun and Saenko, Kate and Wang, Bo},
  booktitle={Proceedings of the IEEE/CVF international conference on computer vision},
  pages={1406--1415},
  year={2019}
}

@inproceedings{resnet,
  title={Deep residual learning for image recognition},
  author={He, Kaiming and Zhang, Xiangyu and Ren, Shaoqing and Sun, Jian},
  booktitle={Proceedings of the IEEE conference on computer vision and pattern recognition},
  pages={770--778},
  year={2016}
}

@inproceedings{coral,
  title={CoRaL: Continual representation learning for overcoming catastrophic forgetting},
  author={Yasar, Mohammad Samin and Iqbal, Tariq},
  booktitle={Proceedings of the 2023 International Conference on Autonomous Agents and Multiagent Systems},
  pages={1969--1978},
  year={2023}
}

@inproceedings{niu2024,
  title={Representation space maintenance: Against forgetting in continual learning},
  author={Niu, Rui and Wu, Zhiyong and Song, Changhe},
  booktitle={2024 International Joint Conference on Neural Networks (IJCNN)},
  pages={1--7},
  year={2024},
  organization={IEEE}
}

@inproceedings{clip,
  title={Continual vision-language representation learning with off-diagonal information},
  author={Ni, Zixuan and Wei, Longhui and Tang, Siliang and Zhuang, Yueting and Tian, Qi},
  booktitle={International Conference on Machine Learning},
  pages={26129--26149},
  year={2023},
  organization={PMLR}
}

@misc{AID,
      title={Activation by Interval-wise Dropout: A Simple Way to Prevent Neural Networks from Plasticity Loss}, 
      author={Sangyeon Park and Isaac Han and Seungwon Oh and Kyung-Joong Kim},
      year={2025},
      eprint={2502.01342},
      archivePrefix={arXiv},
      primaryClass={cs.LG},
      url={https://arxiv.org/abs/2502.01342}, 
}

\end{document}